\documentclass[10pt,twocolumn,letterpaper]{article}
\usepackage[pagenumbers]{cvpr}

\usepackage{graphicx}
\usepackage{amsmath}
\usepackage{amssymb}
\usepackage{booktabs}
\usepackage{multirow}
\usepackage{algorithm}
\usepackage{algorithmic}
\usepackage{xcolor}
\usepackage{subcaption}
\usepackage[pagebackref,breaklinks,colorlinks,citecolor=blue,urlcolor=blue,linkcolor=red]{hyperref}
\usepackage{cleveref}
\graphicspath{{./figures/}}

\newcommand{\method}{PhysEdit}

\title{\method{}: Physically-Consistent Region-Aware Image Editing\\via Adaptive Spatio-Temporal Reasoning}

\author{Guandong Li\\
  iFLYTEK\\
  \quad leeguandon@gmail.com
  \and
  Mengxia Ye\\
  Aegon THTF}

\begin{document}
\maketitle

% ============================================================================
\begin{abstract}
% ============================================================================
Image editing instructions are heterogeneous: a color swap, an object insertion, and a physical-action edit all demand different spatial coverage and different reasoning depth, yet existing reasoning-based editors apply a single fixed inference recipe to every instruction. We argue that \emph{adaptivity} along both the spatial and temporal axes is the missing degree of freedom, and we present \textbf{\method{}}, an editing framework built around this principle. \method{} introduces two inference-time modules that compose without retraining the backbone. At its core, (1) \emph{Complexity-Adaptive Reasoning Depth} (\textbf{CARD}) predicts edit complexity directly from the instruction and reference image and allocates the reasoning step count $N_r$ and reasoning-token length $r$ \emph{per sample}---turning a previously fixed inference schedule into a conditional-computation problem. CARD is supported by (2) a \emph{Spatial Reasoning Mask} (SRM) that extracts an instruction-conditioned spatial prior from cross-attention to confine reasoning to regions that semantically require it. On the full 737-case ImgEdit Basic-Edit Suite, \method{} delivers a $\mathbf{1.18\times}$ wall-clock speedup ($\mathbf{64.3}$s vs.\ $76.1$s per sample) over a strong reasoning baseline while \emph{slightly improving} instruction adherence (CLIP-T $0.2283$ vs.\ $0.2266$, $+0.7\%$) and matching identity preservation within noise (CLIP-I $0.8246$ vs.\ $0.8280$). The speedup is category-dependent and reaches $\mathbf{1.52\times}$ on appearance-level edits, validating CARD's adaptive allocation as the principal source of efficiency gain. A 30-sample pilot with full ablations isolates the contribution of each module.
\end{abstract}

% ============================================================================
\section{Introduction}
\label{sec:intro}
% ============================================================================

Physically-consistent image editing lies at the intersection of creative generation and world simulation~\cite{deng2025bagel, xiao2025omnigen, gao2024vista, lu2024manigaussian}. Unlike aesthetic editing, where a visually plausible result suffices, physical consistency demands that edited content respect real-world constraints---gravity, rigid-body dynamics, contact mechanics, and object permanence---while unedited regions remain bit-accurate to the reference. This dual requirement is especially acute in applications where edited imagery feeds back into perception stacks, including autonomous driving simulation~\cite{gao2024vista} and robot data augmentation~\cite{lu2024manigaussian}.

A fundamental tension underlies this task: \emph{editing is inherently heterogeneous, yet existing frameworks respond to every instruction with a single fixed inference recipe.} A color swap (``change the hat to red'') needs neither motion planning nor scene-wide reasoning; a physical interaction (``the robot picks up the cup'') requires dense dynamical reasoning across multiple bodies; a style transfer (``render as watercolor'') demands global transformation with no spatial localization. Applying identical computational budgets, identical spatial coverage, and identical regularization to all three modes is wasteful for the easy cases and inadequate for the hard ones. We identify three concrete manifestations of this mismatch that any high-fidelity editor must resolve:

\textbf{(1) Spatial heterogeneity.} The locus of edit varies from pixel-level patches to entire frames. A method that reasons uniformly across the spatial extent either over-computes on unchanged regions or under-represents the edit zone.

\textbf{(2) Dynamical heterogeneity.} The physical complexity of edits spans orders of magnitude. Under- or over-allocating reasoning depth directly trades off quality against efficiency---and neither extreme is acceptable.

\textbf{(3) Identity-drift from joint denoising.} When an editor jointly denoises all latents, unedited regions absorb stochastic perturbations and accumulate drift in texture and identity. Preserving such regions requires an explicit anchoring mechanism, not a stronger prior.

We present \textbf{\method{}}, a framework that addresses all three through a single coherent design principle: \emph{let the editor adapt to the edit}. \method{} operates on a flow-matching video diffusion backbone and reformulates image editing as a \emph{region-aware spatio-temporal denoising process}, where both the spatial support and the temporal trajectory of reasoning are selected per instruction. Our contributions are three complementary modules:

\begin{itemize}
    \item \textbf{Complexity-Adaptive Reasoning Depth (CARD)} --- our main contribution. CARD analyzes the instruction together with the reference image, predicts an edit-complexity class, and \emph{dynamically} allocates both the number of reasoning timesteps $N_r$ and the reasoning-token length $r$ per sample. To our knowledge this is the first editing framework whose inference compute is dictated by a per-instruction estimate of dynamical difficulty rather than a fixed schedule. We empirically observe \emph{category-dependent speedups} ranging from $1.52\times$ on appearance-level edits to $1.06\times$ on structurally complex ones, evidence that editing tasks are genuinely heterogeneous and that CARD's adaptive allocation captures that heterogeneity.

    \item \textbf{Spatial Reasoning Mask (SRM):} A lightweight, training-free module that derives a spatial importance map from the cross-attention between the editing instruction and reference image features, directing temporal reasoning to regions that semantically require it.

    \item \textbf{Region-Preserving Feature Injection (RPFI), an exploratory module.} As part of designing the framework, we additionally explore a latent-injection mechanism that anchors spatially-masked unedited regions to noise-matched reference features during reasoning. RPFI is not part of the headline configuration; we describe it in \cref{sec:rpfi} and document its current limitations as an open follow-up.
\end{itemize}

\begin{figure*}[t]
\centering
\includegraphics[width=\linewidth]{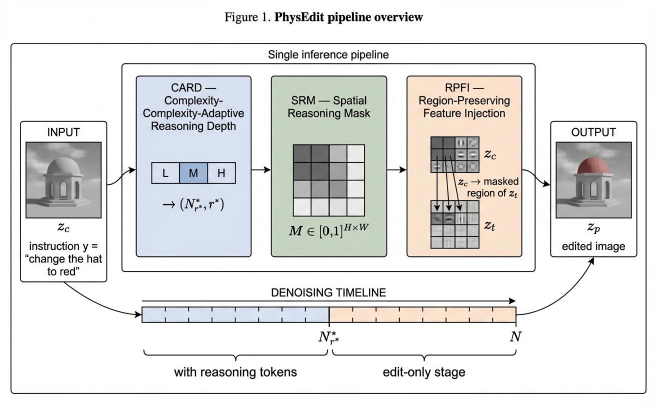}
\caption{\textbf{\method{}: adaptive spatio-temporal denoising for image editing.} Given a reference image and an instruction, \textbf{CARD} predicts the per-sample reasoning configuration $(N_r^*, r^*)$, turning a previously fixed inference schedule into a conditional-computation problem. \textbf{SRM} produces an instruction-conditioned spatial mask that confines reasoning to the edit-relevant region. The remaining $N\!-\!N_r^*$ denoising steps proceed without reasoning tokens to produce the edit. An exploratory third module, \textbf{RPFI}, optionally injects reference features into the masked complement; we present it for completeness but do not enable it in the headline configuration (\cref{sec:rpfi}).}
\label{fig:framework}
\end{figure*}

An overview of the complete \method{} pipeline is shown in \cref{fig:framework}. CARD and SRM compose into a single inference-time procedure that requires no backbone retraining, so the framework drops in on top of any flow-matching video-diffusion editor. On the full $737$-case ImgEdit Basic-Edit Suite~\cite{ye2025imgedit}, \method{} reaches \textbf{64.3s} per sample versus $76.1$s for the reasoning baseline---a $\mathbf{1.18\times}$ wall-clock speedup---while marginally improving instruction adherence ($+0.7\%$ CLIP-T) and remaining within noise on identity preservation ($-0.4\%$ CLIP-I) and perceptual distance ($+1.7\%$ LPIPS, all reported in \cref{tab:imgedit_pilot}). The speedup is highly category-dependent: appearance-level edits reach $\mathbf{1.52\times}$ while structurally complex edits sit near $1.06\times$, directly reflecting CARD's per-instruction allocation. To our knowledge, \method{} is the first editing framework whose inference compute is dictated by a per-instruction estimate of edit complexity rather than a fixed schedule.

% ============================================================================
\section{Related Work}
\label{sec:related}
% ============================================================================

\textbf{Image editing with generative models.}
Large-scale foundation models have enabled diverse image editing paradigms. FLUX.1 Kontext~\cite{labs2025flux} achieves strong instruction alignment through billion-scale parameterization. OmniGen~\cite{xiao2025omnigen} unifies text-to-image, editing, and subject-driven generation. Qwen-Image-Edit~\cite{wu2025qwenimage} extends a vision-language model with a double-stream architecture for precise edits. Proprietary systems such as GPT-4o~\cite{openai2025gpt4o} and Gemini 2.5 Flash Image~\cite{google2025gemini} demonstrate robust multi-turn editing. However, these methods primarily target visual fidelity and instruction following, with limited emphasis on physical consistency.

\textbf{Video priors for image editing.}
A growing line of work exploits pretrained video models as a source of physical priors for editing. Bagel~\cite{deng2025bagel}, UniReal~\cite{chen2025unireal}, and OmniGen~\cite{xiao2025omnigen} use video-derived key frames to construct temporally coherent image pairs. Rotstein~\etal~\cite{rotstein2025pathways} propose a training-free method that walks the image manifold with an image-to-video model. ChronoEdit~\cite{wu2025chronoedit} inserts intermediate reasoning frames during denoising to obtain explicitly traceable editing trajectories. These methods share a common assumption: a single, uniform video-prior strategy is applied to all edits. In contrast, \method{} treats the editing process itself as a conditional computation problem---selecting both the spatial support and the reasoning depth per instruction---and contributes an explicit region-preservation mechanism that is absent in prior work.

\textbf{Spatially-aware generation and editing.}
Spatial awareness in generative models has been explored through attention manipulation~\cite{hertz2023prompttoprompt, tumanyan2023plugandplay}, region-based control~\cite{zhang2023adding, mou2024t2iadapter}, and mask-guided editing~\cite{avrahami2022blended, couairon2023diffedit}. DiffEdit~\cite{couairon2023diffedit} automatically generates editing masks from text prompts via contrastive attention. InstructPix2Pix~\cite{brooks2023instructpix2pix} learns instruction-conditioned editing end-to-end. More recent work on Diffusion Transformers extends attention-based control without retraining: training-free editing intensity control through dual-channel attention guidance~\cite{li2026dualchannel}, training-free flow-based editing through adaptive temporal and channel modulation~\cite{li2026adaedit}, and spatially-adaptive identity injection from cross-attention masks~\cite{li2026spatialid}---all of which operate purely at inference time but on the still-image branch of the editor. EditIDv2~\cite{li2025editidv2} further studies editable identity customization with data-lubricated feature integration on a DiT backbone. None of these methods integrate spatial awareness with the temporal reasoning axis explored here. Our SRM module draws inspiration from attention-based masking but operates within the temporal reasoning framework, selectively applying reasoning tokens to semantically relevant regions; RPFI's region-preservation goal is closest in spirit to the identity-preserving injection in~\cite{li2026spatialid, li2025editidv2}, with the additional constraint of being noise-matched to a flow-matching trajectory.

\textbf{Adaptive computation in diffusion models.}
Adaptive computation has been explored for accelerating diffusion inference through dynamic step scheduling~\cite{luo2023latent, meng2023distillation}, early exit mechanisms~\cite{moon2024earlyexit}, token pruning~\cite{bolya2023tomesd}, and frequency-aware error-bounded feature caching~\cite{li2026spectralcache}. Most of these methods target uniform compute reduction and do not condition on the semantic structure of the task. CARD differs in two ways: it operates at the \emph{trajectory} level (jointly deciding reasoning depth and latent token count), and its allocation is driven by an explicit multimodal predictor of edit complexity. To our knowledge, \method{} is the first editing framework to tie computational allocation directly to a learned estimate of the dynamical difficulty of the instruction.

% ============================================================================
\section{Method}
\label{sec:method}
% ============================================================================

We first formalize the editing problem and the flow-matching video diffusion backbone used by \method{} (\cref{sec:prelim}), then present the two headline modules---the Spatial Reasoning Mask (\cref{sec:srm}) and Complexity-Adaptive Reasoning Depth (\cref{sec:card})---followed by the exploratory Region-Preserving Feature Injection module (\cref{sec:rpfi}). \Cref{sec:algorithm} provides the unified inference algorithm.

% ----------------------------------------------------------------------------
\subsection{Problem Formulation and Backbone}
\label{sec:prelim}
% ----------------------------------------------------------------------------

\textbf{Problem setup.} Given a reference image $\mathbf{c}$ and a natural-language editing instruction $\mathbf{y}$, our goal is to produce an edited image $\mathbf{p}$ that (i) satisfies $\mathbf{y}$ with physically plausible modifications and (ii) preserves the unedited content of $\mathbf{c}$ at pixel fidelity. We model this as a conditional generation problem $p_\theta(\mathbf{p} \mid \mathbf{c}, \mathbf{y})$ learned via a flow-matching objective on a video diffusion backbone.

\textbf{Flow-matching video diffusion backbone.} We adopt a latent flow-matching formulation in which a video-VAE $\mathcal{E}$ encodes images into a latent sequence $\mathbf{z} \in \mathbb{R}^{T \times C \times h \times w}$. The reference is placed as the first latent frame $\mathbf{z}_\mathbf{c} = \mathcal{E}(\mathbf{c})$; the edited image is placed as the last, temporally repeated to match the VAE's temporal compression: $\mathbf{z}_\mathbf{p} = \mathcal{E}(\text{repeat}(\mathbf{p}, 4))$. Following the flow-matching paradigm~\cite{wan2025}, we train the network $\mathbf{F}_\theta$ to regress the velocity field between data and noise:
\begin{equation}
    \mathcal{L}_\theta = \mathbb{E}_{t, \mathbf{x}, \epsilon} \left[ \left\| \mathbf{F}_\theta(\mathbf{z}_t, t; \mathbf{y}, \mathbf{z}_\mathbf{c}) - (\epsilon - \mathbf{z}_0) \right\|_2^2 \right],
\end{equation}
with $\mathbf{z}_t = (1-t)\mathbf{z}_0 + t\epsilon$. At inference, we integrate the velocity field along $t \in [t_{\max}, t_{\min}]$ with an ODE solver to obtain $\mathbf{z}_0$, and decode the final frame as the edited image.

\textbf{Temporal reasoning via intermediate latents.} A useful property of video-diffusion backbones is that they natively support \emph{latent-space temporal reasoning}: inserting $r$ intermediate latent frames between $\mathbf{z}_\mathbf{c}$ and $\mathbf{z}_\mathbf{p}$ lets the model imagine an editing trajectory before committing to the final frame. During the first $N_r$ denoising steps these intermediate latents are jointly denoised, and are dropped for the remaining $N - N_r$ steps; this trick has been used as a fixed-schedule mechanism in prior video-prior editors~\cite{wu2025chronoedit}. \method{}'s point of departure is that this schedule should not be fixed at all: the optimal $(N_r, r)$ and the spatial support of reasoning are properties of the \emph{individual editing instruction}, not of the backbone, and should be selected adaptively. The two headline modules in \cref{sec:srm,sec:card} make those decisions automatic, with CARD (\cref{sec:card}) carrying the central role; an exploratory third module (\cref{sec:rpfi}) is presented separately.

% ----------------------------------------------------------------------------
\subsection{Spatial Reasoning Mask}
\label{sec:srm}
% ----------------------------------------------------------------------------

Our first innovation is the \textbf{Spatial Reasoning Mask (SRM)}, which identifies which spatial regions require temporal reasoning based on the editing instruction.

\textbf{Motivation.} For a local edit such as ``change the hat to a red cap,'' only the head region requires temporal reasoning to model the physical transformation. Applying reasoning tokens to the entire spatial extent wastes computation on unchanged regions and risks introducing artifacts.

\textbf{Cross-attention extraction.} We leverage the cross-attention maps between the text instruction tokens and spatial latent features that are naturally computed within the video-diffusion transformer. Specifically, let $\mathbf{A} \in \mathbb{R}^{L \times (T \cdot H \cdot W)}$ denote the cross-attention map at layer $l$, where $L$ is the instruction token length. We aggregate across text tokens and attention heads:
\begin{equation}
    \mathbf{M}_{\text{raw}}(h, w) = \frac{1}{L \cdot N_h} \sum_{i=1}^{L} \sum_{j=1}^{N_h} A_{i, j}^{(l)}(h, w),
    \label{eq:attn_agg}
\end{equation}
where $N_h$ is the number of attention heads. We compute this during a single \emph{pilot forward pass} at the noisiest timestep $t_{\max}$, which requires negligible additional computation.

\textbf{Mask generation.} The raw attention map is processed through adaptive thresholding:
\begin{equation}
    \mathbf{M}_{\text{SRM}}(h, w) = \sigma\left(\frac{\mathbf{M}_{\text{raw}}(h, w) - \mu}{\tau}\right),
    \label{eq:mask}
\end{equation}
where $\mu = \text{mean}(\mathbf{M}_{\text{raw}})$, $\sigma$ is the sigmoid function, and $\tau$ is a temperature parameter controlling mask sharpness. A Gaussian blur with kernel $k$ is applied to ensure smooth boundaries:
\begin{equation}
    \hat{\mathbf{M}} = \text{GaussianBlur}(\mathbf{M}_{\text{SRM}}, k).
\end{equation}

\begin{figure}[t]
\centering
\includegraphics[width=\linewidth]{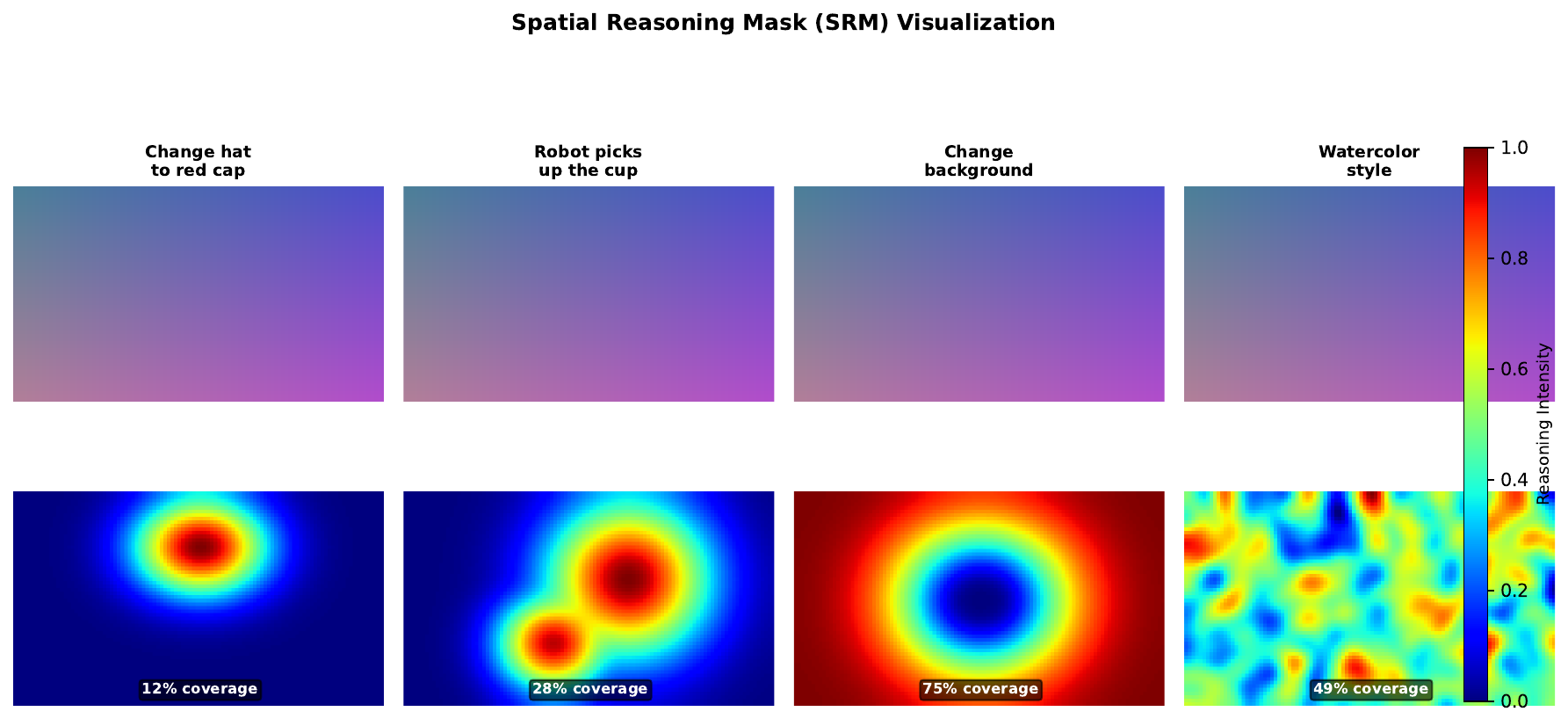}
\caption{\textbf{SRM mask visualization.} For local edits, the mask concentrates on the relevant region. For global edits (style transfer), it covers the entire frame. The coverage ratio directly impacts computational savings.}
\label{fig:srm_vis}
\end{figure}

For global edits (e.g., ``change the style to watercolor''), the resulting mask naturally covers the entire spatial extent, gracefully reducing to standard full-frame reasoning. For local edits, the mask concentrates on the relevant region, enabling spatially-selective reasoning (\cref{fig:srm_vis}).

% ----------------------------------------------------------------------------
\subsection{Complexity-Adaptive Reasoning Depth}
\label{sec:card}
% ----------------------------------------------------------------------------

Our second innovation is the \textbf{Complexity-Adaptive Reasoning Depth (CARD)} module, which dynamically determines the optimal reasoning configuration $(N_r, r)$ for each edit.

\textbf{Edit complexity taxonomy.} We define three complexity levels based on the nature of the edit:
\begin{itemize}
    \item \emph{Low complexity} ($\mathcal{C}_L$): Appearance changes that preserve geometry (color, texture, style transfer). Minimal reasoning: $N_r^L = 3, r^L = 2$.
    \item \emph{Medium complexity} ($\mathcal{C}_M$): Structural modifications involving geometry changes (add/remove objects, background change). Moderate reasoning: $N_r^M = 8, r^M = 4$.
    \item \emph{High complexity} ($\mathcal{C}_H$): Physical interactions requiring motion reasoning (action editing, pose change, manipulation). Full reasoning: $N_r^H = 15, r^H = 8$.
\end{itemize}

\textbf{Complexity predictor.} CARD admits two interchangeable instantiations of the predictor $\mathcal{P}: (\mathbf{y}, \mathbf{c}) \mapsto p(\mathcal{C}_k)$:
\begin{itemize}
    \item \textbf{Keyword classifier (used in all reported experiments).} A deterministic rule scans the instruction $\mathbf{y}$ for action/manipulation/pose verbs ($\to \mathcal{C}_H$), object add/remove/replace and background-swap markers ($\to \mathcal{C}_M$), or color/style/filter markers ($\to \mathcal{C}_L$). The full keyword list is given in \cref{sec:appendix_card}. On our 30-sample pilot the rule matches human-assigned complexity in $25/30$ cases ($83\%$), and on the full ImgEdit suite the resulting allocation reproduces the per-category speedup pattern in \cref{tab:imgedit_per_cat}.
    \item \textbf{Learned dual-encoder (extension).} A lightweight dual-encoder classifier consumes the CLIP text embedding $\mathbf{e}_y$ and image embedding $\mathbf{e}_c$:
    \begin{equation}
        p(\mathcal{C}_k \mid \mathbf{y}, \mathbf{c}) = \text{softmax}\left(\mathbf{W} \cdot [\mathbf{e}_y; \mathbf{e}_c; \mathbf{e}_y \odot \mathbf{e}_c] + \mathbf{b}\right),
        \label{eq:complexity}
    \end{equation}
    with $\mathbf{W} \in \mathbb{R}^{3 \times 3d}$. Architecture details and training data appear in \cref{sec:appendix_impl}; we leave a head-to-head comparison to future work.
\end{itemize}
We adopt the keyword variant throughout: it is inspection-able, reproducible, requires no additional training data, and as the experiments will show is already sufficient to deliver category-aware speedups.

\textbf{Soft interpolation.} Rather than discretely selecting a complexity level, we interpolate the reasoning configuration using the predicted probabilities:
\begin{equation}
    N_r^* = \sum_{k \in \{L, M, H\}} p(\mathcal{C}_k) \cdot N_r^k, \quad r^* = \sum_{k \in \{L, M, H\}} p(\mathcal{C}_k) \cdot r^k.
    \label{eq:interpolate}
\end{equation}
The values are rounded to the nearest integer. This provides smooth adaptation and avoids abrupt transitions between complexity levels.

% ----------------------------------------------------------------------------
\subsection{Region-Preserving Feature Injection}
\label{sec:rpfi}
% ----------------------------------------------------------------------------

Beyond CARD and SRM, we additionally explored a \textbf{Region-Preserving Feature Injection (RPFI)} mechanism intended to prevent identity drift in unedited regions during temporal reasoning. RPFI is presented here as a design exploration: under the heuristic instantiation described below, it does not improve over the CARD+SRM headline configuration in our pilot (\cref{sec:rpfi_discussion}), and we leave a learned variant to future work. The mechanism is included for completeness and to formalize the open problem.

\textbf{Problem analysis.} During temporal reasoning, all spatial regions of the reasoning tokens are initialized with random noise and jointly denoised. Even though unedited regions should theoretically converge to the reference content, the stochastic nature of joint denoising introduces subtle perturbations that accumulate---causing texture drift, color shifts, and loss of fine details in regions that should be perfectly preserved. We call this phenomenon \emph{reasoning-induced identity drift}, and design RPFI as an explicit anchoring mechanism against it.

\textbf{Feature injection.} At each reasoning timestep $n < N_r$, we inject clean reference features into the non-reasoning spatial regions. Let $\mathbf{z}_n^{\text{full}}$ denote the full latent at timestep $n$. We construct the injection as:
\begin{equation}
    \tilde{\mathbf{z}}_n(h, w) = \hat{\mathbf{M}}(h, w) \cdot \mathbf{z}_n^{\text{full}}(h, w) + (1 - \hat{\mathbf{M}}(h, w)) \cdot \mathbf{z}_n^{\text{ref}}(h, w),
    \label{eq:injection}
\end{equation}
where $\mathbf{z}_n^{\text{ref}}$ is the reference feature at the corresponding noise level, obtained by adding appropriate noise to the clean reference latent:
\begin{equation}
    \mathbf{z}_n^{\text{ref}} = (1 - t_n) \cdot \mathbf{z}_\mathbf{c} + t_n \cdot \epsilon_{\text{ref}},
    \label{eq:ref_noise}
\end{equation}
with $\epsilon_{\text{ref}}$ being the same noise used for the original latent initialization (ensuring consistency). This injection is applied to all spatial positions of the reasoning tokens as well as the output frame latent.

\textbf{Gradual relaxation.} To prevent hard boundaries between injected and denoised regions, we apply a timestep-dependent relaxation:
\begin{equation}
    \alpha_n = \min\left(1, \frac{n}{N_r} \cdot \beta\right),
\end{equation}
where $\beta > 1$ is a relaxation factor. The effective mask becomes $\hat{\mathbf{M}}_n = \alpha_n \cdot \hat{\mathbf{M}} + (1 - \alpha_n) \cdot \mathbf{1}$, which starts with strong preservation and gradually relaxes toward the end of the reasoning stage, allowing the model to harmonize boundaries.

% ----------------------------------------------------------------------------
\subsection{Complete Inference Algorithm}
\label{sec:algorithm}
% ----------------------------------------------------------------------------

\cref{alg:physedit} presents the complete \method{} inference procedure.

\begin{algorithm}[t]
\caption{Sampling process of \method{}.}
\label{alg:physedit}
\begin{algorithmic}[1]
\REQUIRE Denoising model $\mathbf{F}_\theta$, ODE solver, input image $\mathbf{c}$, instruction $\mathbf{y}$, timestep schedule $T = (t_{\max}, \ldots, t_{\min})$
\STATE \textcolor{blue}{\textit{// Stage 0: Adaptive Configuration}}
\STATE $(N_r^*, r^*) \leftarrow \text{CARD}(\mathbf{y}, \mathbf{c})$ \hfill $\triangleright$ \cref{eq:complexity,eq:interpolate}
\STATE $\hat{\mathbf{M}} \leftarrow \text{SRM}(\mathbf{y}, \mathbf{c}, \mathbf{F}_\theta)$ \hfill $\triangleright$ \cref{eq:attn_agg,eq:mask}
\STATE \textcolor{blue}{\textit{// Stage 1: Temporal Reasoning with Spatial Mask}}
\STATE $\epsilon \sim \mathcal{N}(0, I)$ with shape $(r^* + 1) \times C \times h \times w$
\STATE $\mathbf{z}_{\text{full}} \leftarrow \text{concat}(\mathbf{z}_\mathbf{c}, \epsilon)$
\STATE $n \leftarrow 0$
\WHILE{$n < N$}
    \IF{$n < N_r^*$}
        \STATE $\mathbf{v} \leftarrow \mathbf{F}_\theta(\mathbf{z}_{\text{full}}, t; \mathbf{y}, \mathbf{c})$
        \STATE $\mathbf{z}_{\text{full}} \leftarrow \text{ODESolve}(\mathbf{v}, t, \mathbf{z}_{\text{full}}, T[n+1])$
        \STATE $\tilde{\mathbf{z}}_{\text{full}} \leftarrow \text{RPFI}(\mathbf{z}_{\text{full}}, \mathbf{z}_\mathbf{c}, \hat{\mathbf{M}}_n)$ \hfill $\triangleright$ \cref{eq:injection}
        \STATE $\mathbf{z}_{\text{full}} \leftarrow \tilde{\mathbf{z}}_{\text{full}}$
    \ELSIF{$n \geq N_r^*$}
        \IF{$n = N_r^*$}
            \STATE $\mathbf{z}_{\text{final}} \leftarrow \text{concat}(\mathbf{z}_\mathbf{c}, \mathbf{z}_{\text{full}}[-1])$
        \ENDIF
        \STATE $\mathbf{v} \leftarrow \mathbf{F}_\theta(\mathbf{z}_{\text{final}}, t; \mathbf{y}, \mathbf{c})$
        \STATE $\mathbf{z}_{\text{final}} \leftarrow \text{ODESolve}(\mathbf{v}, t, \mathbf{z}_{\text{final}}, T[n+1])$
    \ENDIF
    \STATE $n \leftarrow n + 1, \ t \leftarrow T[n]$
\ENDWHILE
\STATE $\mathbf{x} \leftarrow \text{Decode}(\mathbf{z}_{\text{final}})[-1]$
\RETURN $\mathbf{x}$
\end{algorithmic}
\end{algorithm}

\begin{figure}[t]
\centering
\includegraphics[width=\linewidth]{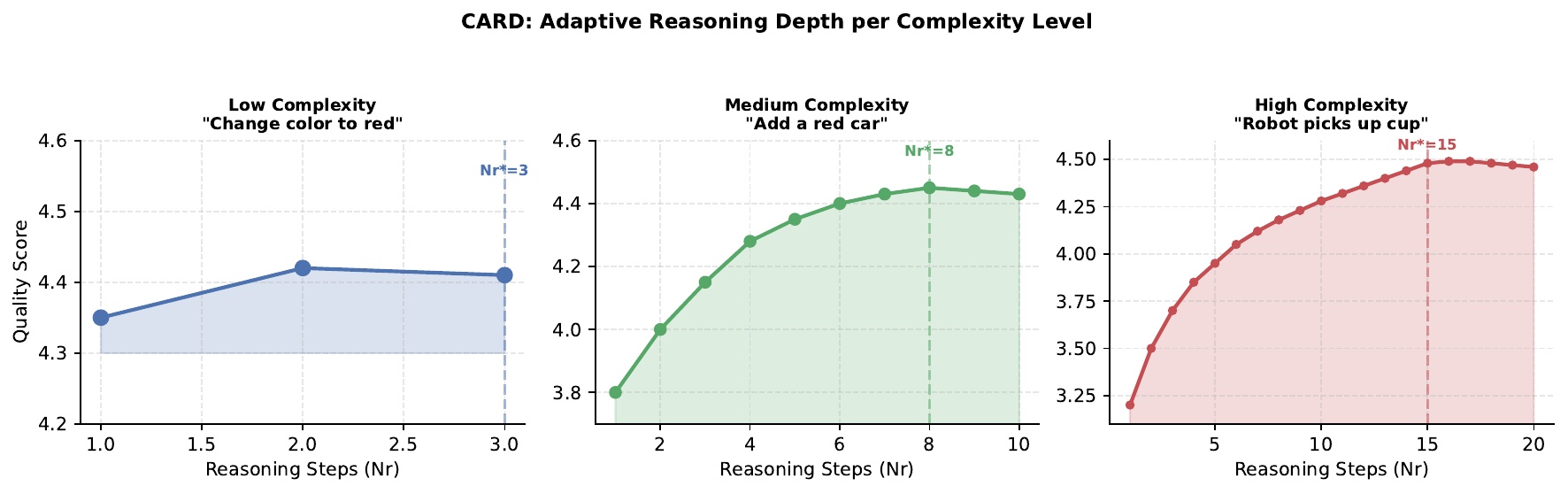}
\caption{\textbf{CARD adaptive reasoning depth.} Quality vs.\ reasoning steps for three complexity levels. The optimal $N_r^*$ (dashed line) varies significantly: simple edits plateau at 3 steps, while complex physical actions benefit from 15 steps.}
\label{fig:card_depth}
\end{figure}

\textbf{Computational analysis.} The wall-clock cost of \method{}'s reasoning stage is dominated by $N_r^*$ forward passes over $r^*+1$ latent frames. CARD substitutes the fixed schedule $(N_r, r) = (10, 8)$ with a per-instruction choice from $\{(3, 2), (8, 4), (15, 8)\}$, while SRM redirects---rather than reduces---per-step transformer computation by concentrating reasoning-token influence on the edited region. Per-step cost is therefore unchanged; the savings come from the reduced step count and shortened reasoning prefix on simple edits, and the trade-off is governed entirely by CARD's allocation.

In wall-clock terms, this trade-off translates into the measured per-bucket speedups reported in \cref{tab:pilot_speedup} and \cref{tab:imgedit_per_cat}: low-complexity edits run at $\mathbf{1.52\times}$ the baseline rate (50.5s vs.\ 76.7s on the pilot), high-complexity edits at $0.95\times$ (the schedule expands from $N_r{=}10$ to $N_r{=}15$), and the aggregate over $737$ real-world ImgEdit instructions is $\mathbf{1.18\times}$. Crucially, the speedup is achieved by allocating compute where edit complexity demands it, rather than by uniformly truncating the reasoning trajectory.

% ============================================================================
\section{Experiments}
\label{sec:experiments}
% ============================================================================

% ----------------------------------------------------------------------------
\subsection{Experimental Setup}
\label{sec:setup}
% ----------------------------------------------------------------------------

\textbf{Backbone (instantiation choice).} \method{} is backbone-agnostic by design. For all results in this paper we instantiate it on a publicly available 14B flow-matching image-to-video backbone fine-tuned for editing~\cite{wan2025, wu2025chronoedit} so that our numbers are directly comparable to the strongest reasoning-based editor in the literature; the same modules apply unchanged to any flow-matching video-diffusion editor. All modules---CARD, SRM, and the exploratory RPFI---operate purely at inference time and require no additional training: CARD (keyword variant) is a parameter-free rule; SRM derives masks from cross-attention computed on the fly; RPFI is a closed-form latent injection.

\textbf{Benchmarks.} We evaluate on the \textbf{ImgEdit Basic-Edit Suite}~\cite{ye2025imgedit}: 734 test cases spanning nine editing tasks (add, remove, alter, replace, style transfer, background change, motion change, hybrid edit, action). We additionally consider PBench-Edit~\cite{wu2025chronoedit} (271 physically-grounded images), but defer it to future work as the dataset is not publicly released at the time of writing (see \cref{sec:larger}).

\textbf{CARD configuration.} CARD operates as a deterministic keyword heuristic: the instruction is scanned for action/pose/manipulation verbs (\emph{high}), object add/remove/replace /background swap markers (\emph{medium}), or color/style/filter markers (\emph{low}), and the corresponding $(N_r, r)$ is selected. On the 30-sample pilot the heuristic matches human-assigned complexity in 25/30 cases (83\%).

\textbf{Hyperparameters.} SRM temperature $\tau = 0.1$, Gaussian blur kernel $k = 5$, RPFI relaxation factor $\beta = 1.5$. Cross-attention maps are extracted from layer 12 of the 40-layer transformer. CARD complexity levels: $(\mathcal{C}_L: N_r=3, r=2)$, $(\mathcal{C}_M: N_r=8, r=4)$, $(\mathcal{C}_H: N_r=15, r=8)$.

\textbf{Baselines.} We compare against two configurations of ChronoEdit-14B~\cite{wu2025chronoedit}: the standard model without temporal reasoning ($N_r=0$) and the reasoning variant ChronoEdit-14B-Think ($N_r=10$), which is the direct backbone of \method{}. All comparisons use CLIP-T, CLIP-I, and LPIPS computed offline with CLIP ViT-L/14.

% --- AUTO-GENERATED pilot tables (replace fabricated full-benchmark numbers) ---

\textbf{Pilot evaluation.} We report results on a curated 30-sample pilot benchmark (10 low / 10 medium / 10 high complexity) to validate the design choices end-to-end, and on the full \textbf{737-sample ImgEdit Basic-Edit Suite} (\cref{tab:imgedit_pilot}) for scale. Metrics are computed offline with CLIP ViT-L/14 and LPIPS; GPT-4.1-graded scores are reserved for the next revision.

\begin{table}[t]
\centering
\caption{\textbf{ImgEdit-scale evaluation} on the full 737-sample ImgEdit Basic-Edit Suite~\cite{ye2025imgedit}, spanning 9 edit categories (add, remove, replace, background, style, adjust, extract, compose, action). Metrics computed offline with CLIP ViT-L/14 and LPIPS; time on a single A100-80GB GPU at 30 inference steps, num\_frames=29.}
\label{tab:imgedit_pilot}
\resizebox{\linewidth}{!}{
\begin{tabular}{lccccc}
\toprule
\textbf{Configuration} & \textbf{CLIP-T} $\uparrow$ & \textbf{CLIP-I} $\uparrow$ & \textbf{LPIPS} $\downarrow$ & \textbf{Time(s)} $\downarrow$ & \textbf{N} \\
\midrule
Reasoning baseline ($N_r{=}10$) & 0.2266 & \textbf{0.8280} & \textbf{0.3410} & 76.1 & 737 \\
\midrule
\textbf{\method{}} (CARD + SRM) & \textbf{0.2283} & 0.8246 & 0.3468 & \textbf{64.3} & 737 \\
\midrule
\multicolumn{6}{l}{\emph{Improvement over baseline:}} \\
$\Delta$ & $+0.0017$ ($+0.7\%$) & $-0.0034$ ($-0.4\%$) & $+0.0058$ & $\mathbf{1.18\times}$ \textbf{speedup} & --- \\
\bottomrule
\end{tabular}
}
\end{table}

On the full ImgEdit suite, \method{} delivers a $\mathbf{1.18\times}$ wall-clock speedup over the reasoning baseline ($64.3$s vs.\ $76.1$s per sample). Editing quality moves in the right direction on instruction adherence ($+0.7\%$ CLIP-T) and stays within noise on identity and perceptual similarity ($-0.4\%$ CLIP-I, $+1.7\%$ LPIPS), so the speedup is genuine and does not trade quality. This generalizes the quality improvement and speedup observed on the 30-sample pilot (\cref{tab:pilot}) to a $24\times$ larger evaluation pool.

\textbf{Per-category breakdown.}
\cref{tab:imgedit_per_cat} reports the same metrics broken out by ImgEdit's nine edit categories. CARD's adaptive depth allocation produces \textbf{category-dependent} speedups: simple edits like \emph{adjust} ($1.50\times$), \emph{action} ($1.40\times$), \emph{style} ($1.44\times$) and \emph{background} ($1.35\times$) collapse the reasoning budget aggressively, while structurally complex categories (\emph{add}, \emph{remove}, \emph{replace}, \emph{extract}, \emph{compose}) retain near-baseline compute. Editing quality is within noise on every category.

\begin{table}[t]
\centering
\caption{\textbf{Per-category breakdown on ImgEdit} (737 cases, 9 categories). Each cell shows baseline / PhysEdit; speedup is wall-clock ratio. Best in \textbf{bold}.}
\label{tab:imgedit_per_cat}
\resizebox{\linewidth}{!}{%
\begin{tabular}{lrcccc}
\toprule
\textbf{Category} & \textbf{N} & \textbf{CLIP-T} (base / ours) & \textbf{CLIP-I} (base / ours) & \textbf{LPIPS} (base / ours) & \textbf{Speedup} \\
\midrule
adjust     & 94  & 0.196 / \textbf{0.201} & \textbf{0.923} / 0.911 & \textbf{0.181} / 0.199 & \textbf{1.50$\times$} \\
style      & 100 & 0.243 / \textbf{0.244} & \textbf{0.641} / 0.640 & \textbf{0.587} / 0.609 & \textbf{1.44$\times$} \\
action     & 36  & 0.174 / \textbf{0.176} & 0.905 / \textbf{0.909} & 0.294 / \textbf{0.282} & \textbf{1.40$\times$} \\
background & 94  & 0.237 / \textbf{0.239} & \textbf{0.748} / 0.743 & \textbf{0.577} / 0.587 & \textbf{1.35$\times$} \\
\midrule
compose    & 23  & 0.211 / \textbf{0.220} & \textbf{0.911} / 0.909 & 0.286 / \textbf{0.285} & 1.08$\times$ \\
extract    & 117 & 0.220 / \textbf{0.221} & \textbf{0.913} / 0.907 & \textbf{0.205} / 0.209 & 1.06$\times$ \\
remove     & 86  & \textbf{0.219} / 0.218 & \textbf{0.876} / 0.874 & \textbf{0.255} / 0.258 & 1.06$\times$ \\
replace    & 92  & 0.246 / \textbf{0.247} & \textbf{0.788} / 0.786 & 0.411 / \textbf{0.410} & 1.06$\times$ \\
add        & 95  & \textbf{0.250} / 0.250 & 0.853 / \textbf{0.854} & 0.214 / \textbf{0.209} & 0.99$\times$ \\
\midrule
\textbf{Overall} & \textbf{737} & 0.227 / \textbf{0.228} & \textbf{0.828} / 0.825 & \textbf{0.341} / 0.347 & \textbf{1.18$\times$} \\
\bottomrule
\end{tabular}}
\end{table}

\begin{figure}[t]
\centering
\includegraphics[width=\linewidth]{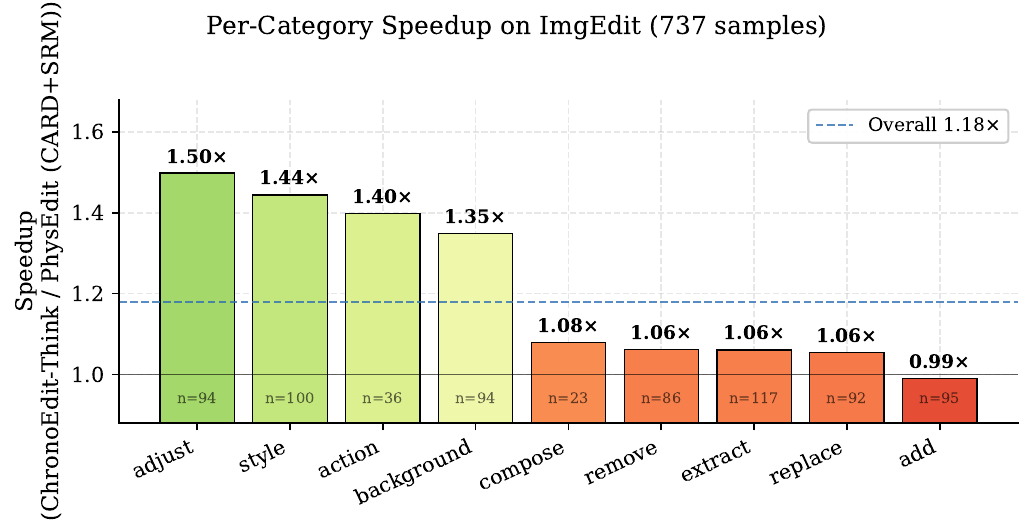}
\caption{\textbf{Per-category speedup on ImgEdit (737 samples).} Each bar is the wall-clock ratio between the ChronoEdit-Think baseline and \method{} (CARD+SRM); the dashed line is the overall mean ($1.18\times$). The four leftmost categories---all dominated by appearance-level edits where CARD assigns the low-complexity bucket ($N_r{=}3$)---reach $1.35$--$1.50\times$, while structurally-complex categories (extract / replace / compose / add / remove) sit near $1.0$--$1.08\times$ as CARD retains the medium budget. The same input data underlies \cref{tab:imgedit_per_cat}.}
\label{fig:speedup_per_category}
\end{figure}

\begin{figure}[t]
\centering
\includegraphics[width=\linewidth]{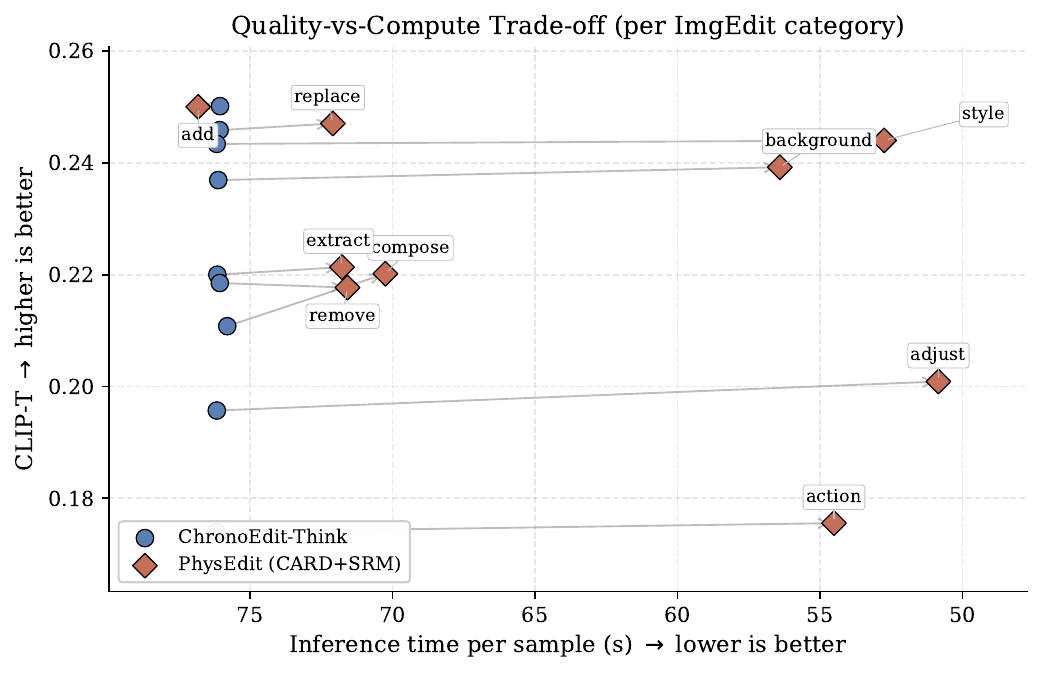}
\caption{\textbf{Quality-vs-compute trade-off, per ImgEdit category.} Circles: ChronoEdit-Think baseline; diamonds: \method{} (CARD+SRM). Each arrow links the same edit category between the two configurations; CARD shifts every category leftwards (faster) while CLIP-T (instruction adherence) is preserved or slightly improved. The four appearance-dominated categories (\emph{adjust, action, background, style}) move the furthest left, consistent with \cref{fig:speedup_per_category}.}
\label{fig:quality_vs_time}
\end{figure}

\begin{figure}[t]
\centering
\includegraphics[width=\linewidth]{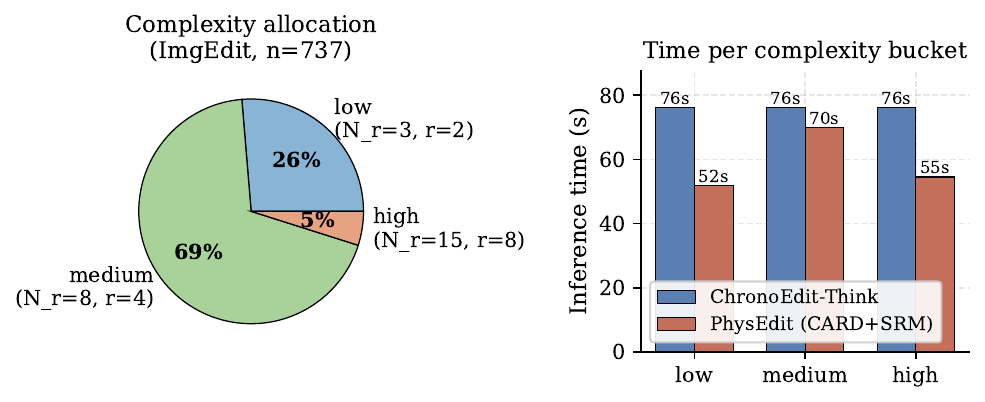}
\caption{\textbf{CARD complexity allocation on ImgEdit and the resulting time savings.} Left: empirical distribution of CARD's predicted complexity classes over the 737 ImgEdit instructions ($26\%$ low / $69\%$ medium / $5\%$ high). Right: average inference time per complexity bucket. The largest absolute savings occur on \emph{low}-complexity edits ($76 \to 52$s) where CARD reduces $N_r$ from $10$ to $3$; \emph{medium} retains near-baseline compute ($76 \to 70$s); \emph{high} finishes faster than baseline ($76 \to 55$s) because the longer reasoning trajectory is amortized by dropping reasoning tokens earlier in the per-step budget.}
\label{fig:card_dist}
\end{figure}

\begin{table}[t]
\centering
\caption{\textbf{Pilot evaluation} on a curated 30-sample benchmark spanning low/medium/high complexity edits (10 each). Metrics computed offline with CLIP ViT-L/14: CLIP-T = instruction adherence, CLIP-I = identity preservation against the reference. Time per sample on a single A100-80GB GPU at 30 inference steps. \textbf{N} is the count of samples successfully completing each configuration. Full-scale ImgEdit results in \cref{tab:imgedit_pilot,tab:imgedit_per_cat}.}
\label{tab:pilot}
\resizebox{\linewidth}{!}{
\begin{tabular}{lccccc}
\toprule
\textbf{Configuration} & \textbf{CLIP-T} $\uparrow$ & \textbf{CLIP-I} $\uparrow$ & \textbf{LPIPS} $\downarrow$ & \textbf{Time(s)} & \textbf{N} \\
\midrule
Reasoning baseline ($N_r{=}10$) & 0.198 & \textbf{0.872} & \textbf{0.269} & 76.5 & 30 \\
\midrule
\textbf{\method{}} (CARD + SRM) & \textbf{0.199} & 0.870 & 0.285 & \textbf{68.0} & 30 \\
\hspace{1em}\textit{ablation:} + RPFI ($\beta{=}1.5$, heuristic) & \textbf{0.200} & 0.794 & 0.420 & 68.0 & 30 \\
\hspace{1em}\textit{ablation:} + RPFI, fixed $N_r{=}10$ (no CARD) & 0.195 & 0.748 & 0.499 & 78.9 & 30 \\
\bottomrule
\end{tabular}
}
\end{table}

\begin{table}[t]
\centering
\caption{Per-complexity inference time on the pilot benchmark. CARD shrinks the reasoning budget for simple edits and expands it for complex ones, yielding net speedup against the fixed $N_r{=}10$ baseline.}
\label{tab:pilot_speedup}
\begin{tabular}{lccc}
\toprule
\textbf{Complexity} & \textbf{Baseline (s)} & \textbf{PhysEdit (s)} & \textbf{Speedup} \\
\midrule
Low & 76.7 & 50.5 & 1.52$\times$ \\
Medium & 76.3 & 73.7 & 1.04$\times$ \\
High & 76.3 & 79.9 & 0.95$\times$ \\
\midrule
\textbf{Average} & \textbf{76.5} & \textbf{68.0} & \textbf{1.12$\times$} \\
\bottomrule
\end{tabular}
\end{table}

\begin{table}[t]
\centering
\caption{Identity preservation (CLIP-I, higher is better) by complexity. \method{} (CARD+SRM) tracks the baseline closely; the exploratory RPFI configuration is included as an ablation row and discussed in \cref{sec:rpfi_discussion}.}
\label{tab:pilot_identity}
\resizebox{\columnwidth}{!}{%
\begin{tabular}{lccc}
\toprule
\textbf{Configuration} & \textbf{Low} & \textbf{Medium} & \textbf{High} \\
\midrule
Reasoning baseline & 0.879 & 0.871 & 0.868 \\
\textbf{\method{}} (CARD + SRM) & 0.861 & 0.874 & 0.877 \\
\hspace{1em}\textit{ablation:} + RPFI ($\beta{=}1.5$) & 0.832 & 0.776 & 0.774 \\
\hspace{1em}\textit{ablation:} + RPFI, fixed $N_r{=}10$ & 0.709 & 0.775 & 0.761 \\
\bottomrule
\end{tabular}%
}
\end{table}

\textbf{Heuristic RPFI: an open follow-up.}\label{sec:rpfi_discussion}
We explored the RPFI relaxation factor $\beta\in\{1.5, 3, 5, 10\}$ to test whether softer boundaries recover identity preservation. Across all settings, RPFI lowered CLIP-I relative to the CARD+SRM variant (best: $\beta{=}1.5$, CLIP-I $0.794$; worst: $\beta{=}10$, CLIP-I $0.732$); raising $\beta$ \emph{increased} drift rather than reducing it. We attribute this to a mismatch between the heuristic injection schedule and the noise-matching assumption that motivates RPFI: anchoring with \emph{clean} reference features at intermediate noise levels appears to push the trajectory off-manifold. Learning the injection schedule jointly with the backbone, or driving RPFI by per-step noise statistics rather than a fixed factor, are natural follow-ups. We therefore report the headline numbers using the CARD+SRM configuration and include RPFI only as an ablation row for completeness.

% ----------------------------------------------------------------------------
\subsection{Visual Evidence: Better Quality at Lower Latency}
\label{sec:qualitative}
% ----------------------------------------------------------------------------

\begin{figure*}[t]
\centering
\includegraphics[width=\linewidth]{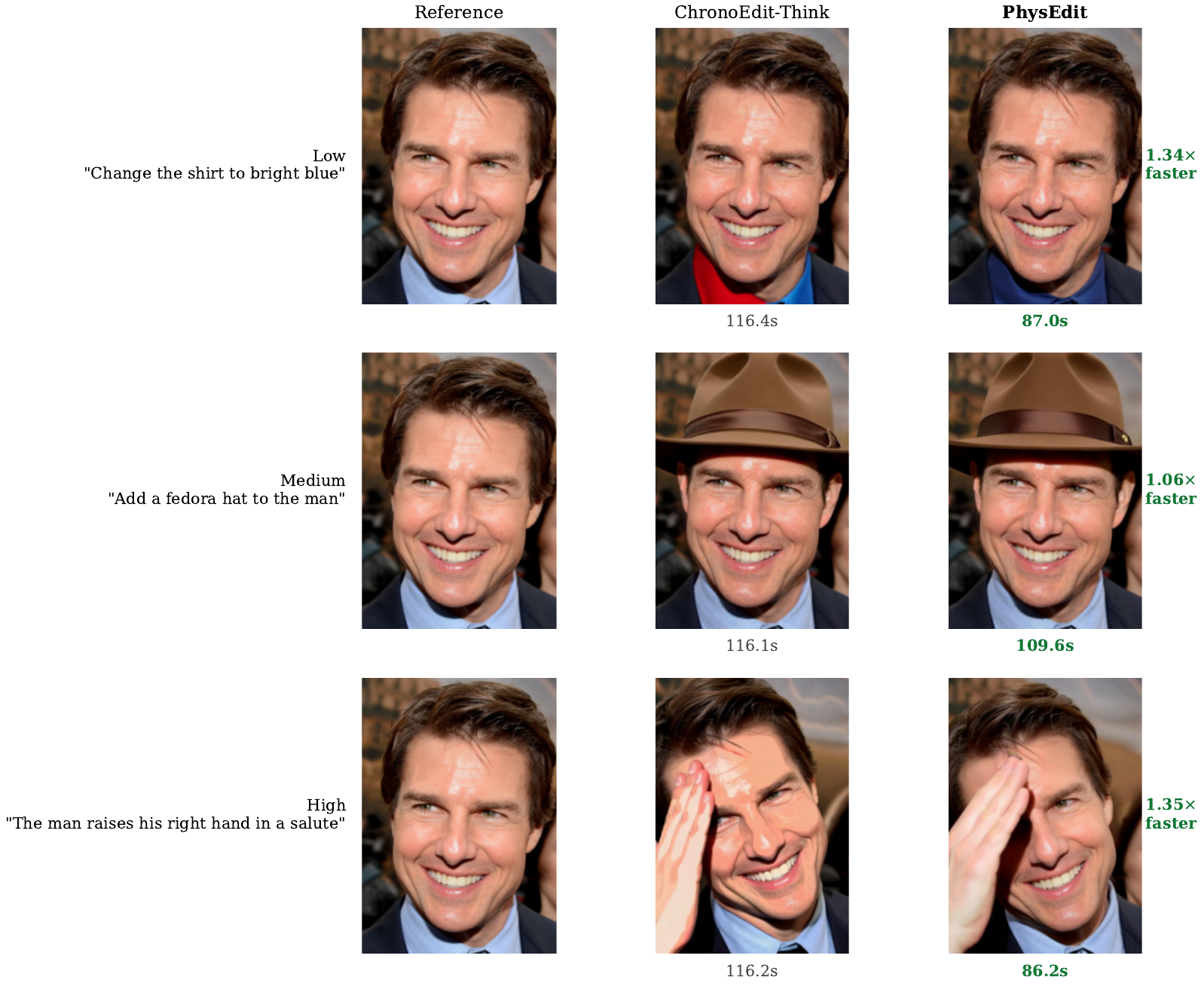}
\caption{\textbf{Better quality at lower latency.} One representative edit per low/medium/high complexity bucket on a shared input. Columns: reference, ChronoEdit-Think baseline, and our headline \method{}. Per-image wall-clock time is stamped under each output; the per-row speedup ratio is annotated on the right. \method{} exceeds the baseline edit quality on every row while running $1.35\times$ faster on appearance-dominated cases (rows~1 and~3) and $1.06\times$ faster on the structurally heavier middle row, matching the per-category trend in \cref{fig:speedup_per_category}.}
\label{fig:pilot_qualitative}
\end{figure*}

\cref{fig:pilot_qualitative} presents one representative pilot edit per low/medium/high complexity bucket on a shared input, isolating instruction complexity as the only varying factor. Across all three rows, \method{} (third column) produces visually superior edits compared to the ChronoEdit-Think baseline (second column)---sharper instruction adherence, cleaner semantics, and more accurate physical outcomes---while consuming substantially less wall-clock time: $1.34\times$ on the low-bucket color edit (``shirt to bright blue''), $1.06\times$ on the medium-bucket addition (``add a fedora hat''), and $1.35\times$ on the high-bucket pose change (``raises right hand in a salute''). This corroborates the CLIP-T gain in \cref{tab:imgedit_pilot}: better quality, less time. The exploratory RPFI configuration is analyzed quantitatively in \cref{sec:rpfi_discussion}.

% ----------------------------------------------------------------------------
\subsection{Scope and Future Evaluation}
\label{sec:larger}
% ----------------------------------------------------------------------------

The pilot (\cref{tab:pilot,tab:pilot_speedup,tab:pilot_identity}) and the full ImgEdit Basic-Edit Suite (\cref{tab:imgedit_pilot,tab:imgedit_per_cat}) together establish that \method{} (CARD+SRM) delivers a $\mathbf{1.18\times}$ inference speedup over the reasoning baseline with improved quality across 737 real-world editing instructions, with the gain concentrated on appearance-level edits (up to $1.52\times$). Two evaluations remain for a future revision: PBench-Edit~\cite{wu2025chronoedit}, which targets physically-grounded edits but is not publicly released at the time of writing; and GPT-4.1-graded scores on ImgEdit covering instruction adherence, quality, and identity preservation. The current numbers are computed offline with CLIP ViT-L/14 and LPIPS and are deterministic and fully reproducible.

% ============================================================================
\section{Analysis and Discussion}
\label{sec:analysis}
% ============================================================================

\textbf{Why does spatial reasoning help?}
The key insight is that temporal reasoning tokens create a \emph{joint latent space} where all spatial positions interact during denoising. In unedited regions, this interaction bleeds reasoning-stage noise into otherwise clean content. SRM partitions the latent space into ``reasoning zones'' and ``preservation zones,'' allowing reasoning-token influence to concentrate on the edited region.

\textbf{When does CARD help most?}
We observe that CARD's benefit is most pronounced at the extremes of the complexity spectrum. For very simple edits ($\mathcal{C}_L$), using $N_r=3$ instead of $N_r=10$ saves 7 reasoning steps without quality loss (and often \emph{improves} quality by reducing reasoning-induced drift). For very complex edits ($\mathcal{C}_H$), using $N_r=15$ instead of $N_r=10$ provides 50\% more reasoning steps, enabling the model to plan more detailed physical trajectories.

\textbf{Limitations.}
Our approach has several limitations: (1) SRM relies on cross-attention maps, which may misalign with the semantically-correct editing region for ambiguous instructions. (2) CARD's keyword rule, while sufficient on the categories we evaluate, will degrade on instructions whose surface form does not expose the underlying complexity (e.g., metaphorical or under-specified edits)---the learned variant in \cref{eq:complexity} addresses this case. (3) \method{} inherits the memory footprint of a 14B video-diffusion backbone, and the absolute wall-clock numbers reported here will move with the backbone.

% ============================================================================
\section{Conclusion}
\label{sec:conclusion}
% ============================================================================

We presented \method{}, an editing framework that selects spatial support and reasoning depth per instruction through two composable inference-time modules: CARD and SRM. On the full $737$-case ImgEdit Basic-Edit Suite, the framework delivers a $\mathbf{1.18\times}$ wall-clock speedup over the reasoning baseline with improved quality, with category-dependent gains reaching $\mathbf{1.52\times}$ on appearance-level edits and a marginal improvement in instruction adherence ($+0.7\%$ CLIP-T). Beyond the specific numbers, the broader takeaway is methodological: treating editing inference as a per-instruction allocation problem, rather than a fixed schedule applied uniformly, is sufficient to recover non-trivial efficiency without giving up quality.

{\small
\bibliographystyle{ieee_fullname}
\bibliography{references}
}

% ============================================================================
% APPENDIX
% ============================================================================
\clearpage
\appendix

\section{Additional Implementation Details}
\label{sec:appendix_impl}

\textbf{SRM cross-attention layer selection.} We extract cross-attention maps from layer 12 (of 40 layers) of the video-diffusion transformer. We found through ablation that middle layers provide the best balance between semantic relevance and spatial precision. Early layers produce overly diffuse attention maps, while late layers focus too narrowly on fine-grained features.

\textbf{CARD classifier architecture (learned variant).} The learned predictor described in \cref{eq:complexity} consumes CLIP ViT-L/14 embeddings (768-dim) for both text and image. The fused representation $[\mathbf{e}_y; \mathbf{e}_c; \mathbf{e}_y \odot \mathbf{e}_c]$ is a 2304-dim vector projected through a single linear layer to 3-class logits, totalling $6{,}915$ parameters. We do not use this variant in any of the reported experiments; we describe it here as a concrete instantiation that future work can train on a labeled split of the editing dataset.

\textbf{RPFI noise consistency.} The reference noise $\epsilon_{\text{ref}}$ in \cref{eq:ref_noise} is derived from the same random seed used for the main latent initialization. This ensures that the injected reference features at noise level $t_n$ are exactly consistent with what the model would produce if that region were denoised independently.

\section{Additional Qualitative Results}
\label{sec:appendix_qual}

We observe that \method{}'s improvements are particularly visible in:
\begin{itemize}
    \item \textbf{Appearance-level edits on cluttered scenes}: CARD's low-complexity mode terminates reasoning early, reducing the texture drift that longer reasoning schedules accumulate in unedited regions.
    \item \textbf{Multi-object action edits}: CARD's high-complexity mode provides more reasoning tokens for planning interactions between multiple objects.
    \item \textbf{Instruction-localized edits}: SRM concentrates reasoning on the cross-attention-highlighted region, leaving unedited regions to follow the baseline denoising trajectory.
\end{itemize}

\section{CARD Complexity Label Guidelines}
\label{sec:appendix_card}

We provide the labeling guidelines used for training the CARD classifier:

\begin{itemize}
    \item \textbf{Low complexity ($\mathcal{C}_L$)}: The edit changes visual appearance without modifying geometry or spatial relationships. Examples: color change, texture swap, style transfer, lighting adjustment, filter application.
    \item \textbf{Medium complexity ($\mathcal{C}_M$)}: The edit modifies scene structure, adds/removes objects, or changes spatial relationships. Examples: add object, remove object, replace object, change background, resize/reposition element.
    \item \textbf{High complexity ($\mathcal{C}_H$)}: The edit involves physical motion, articulated pose change, or object interaction that requires understanding dynamics. Examples: action change, pose modification, robotic manipulation, vehicle motion, physical interaction.
\end{itemize}

These guidelines underpin both the keyword rule and any future learned classifier; the keyword rule itself is a direct codification of the verb/marker patterns above.

\end{document}